\documentclass{article}

 \usepackage[dblblindworkshop, final]{neurips_2026}

\usepackage[utf8]{inputenc} 
\usepackage[T1]{fontenc}    
\usepackage{hyperref}       
\usepackage{url}            
\usepackage{booktabs}       
\usepackage{amsfonts}       
\usepackage{nicefrac}       
\usepackage{microtype}      
\usepackage{xcolor}         

\usepackage{amsmath}
\usepackage{amssymb}
\usepackage{graphicx}

\title{ROCOCO: Robust Outline Completion of Occluded Curvilinear Objects}

\author{%
  Ken Zheng \qquad Jaimyn Drake \\
  Electrical Engineering \& Computer Sciences (EECS) \\
  University of California, Berkeley \\
  \texttt{\{kenzhengjk, jaimyndrake\}@berkeley.edu} \\
}

\begin{document}
\maketitle

\begin{abstract}
We propose a learning-based generative model for completing curvilinear features under 
occlusion by environmental obstacles. To this end, we adopt a flow matching-based approach 
that supports the inpainting of occluded regions with the appropriate curvilinear features. 
To eliminate the reliance on privileged information about the occlusion a priori, we also train 
a UNet-based mask prediction model, which can identify regions for inpainting given only the 
occluded observation. The components of our approach are validated via experiments using 
procedurally-generated test cases, and the complete pipeline is validated using large public 
datasets for curvilinear objects. Against off-the-shelf inpainting baselines, ROCOCO is the
strongest method by a wide margin when no occlusion mask is given, and is competitive with
mask-supervised inpainting when one is, achieving best on full-image error and second only to LaMa on
the occluded region, while using a completer of $7.7$M parameters against LaMa's $51$M and
Stable Diffusion's $860$M. We release our \href{https://github.com/kenzhengjk/ROCOCO}{code}.
\end{abstract}

\section{Introduction}
\label{sec:intro}
The perception of curvilinear objects is an important task in robotics, contributing to downstream applications such as the manipulation of deformable linear objects (DLOs) \cite{azulay2025motorcycle}, which promise automation in the electronics, automotive, and manufacturing industries. Such manipulation is challenging in part because of pose estimation difficulties, including self-occlusion. Occlusions are particularly problematic due to the natural variability of segment states in curvilinear objects, unlike rigid-body counterparts. This work therefore addresses occlusion in curvilinear features by producing a learned generative model that completes such features across occlusion-induced discontinuities.

Prior work has investigated tracing curvilinear objects \cite{viswanath2023handloom}, but largely focuses on un-occluded objects fully visible on planar surfaces. More recently, work has addressed occluding objects that produce discontinuities \cite{yuhandloom}, but does so through interactive perception, using a robotic manipulator to remove obstacles and reduce uncertainty, rather than addressing unavoidable occlusions. Other works have used diffusion-based methods to estimate the shape and appearance of partially-occluded objects \cite{ozguroglu2024pix2gestalt}, but primarily consider rigid or piecewise-rigid objects rather than deformable and/or curvilinear ones.

In our work, we employ a similar flow-matching approach \cite{lipman2022flow} to estimate partially-occluded objects, focusing specifically on curvilinear features. Our primary contributions are:
\begin{enumerate}
\item A two-stage learning-based pipeline for completing curvilinear objects under occlusions.
\item Experiments ablating our approach and comparing it with relevant baselines.
\item Results on realistic curvilinear-object datasets showing that our approach is the strongest available method when no occlusion mask is provided, and competitive with mask-supervised inpainting when one is provided, while using a small fraction of the parameters.
\end{enumerate}

\section{Related Work}
\label{sec:related}

\paragraph{Perception of Deformable Linear Objects.} 
When it comes to the perception of deformable linear objects (DLOs), most general image segmentation methods (such as Segment Anything Model \cite{kirillov2023segment} and its descendants) struggle due to the relatively low pixel density of such objects relative to their occupancy area. As a result, bounding-box based methods are more likely to perceive additional objects in the background rather than the sparse curvilinear target. Point prompt-based methods can also struggle with curvilinear objects, as the extended traversal length needed to generate masks from a seed point can conflict with biases towards compact objects. Long DLOs may often be identified as a series of multiple discrete objects, rather than one unified trace.

These weaknesses of general perception methods motivate the development of custom strategies more suited to the form factor of DLOs. While classical methods akin to tint fill \cite{smith1979tint} (also known as flood fill) can be effective for this type of segmentation by gradually traversing across adjacent pixels with similar color values, more sophisticated deep learning methods can further leverage the properties of curvilinear objects by modeling estimation as an autoregressive prediction problem \cite{viswanath2023handloom}. However, these methods typically assume full visibility of the DLO in question, and assume the provision of an initial seed point from which to expand. Some recent work aims to address the challenge of occlusion \cite{yuhandloom}, but does so via physical intervention. In contrast, our work treats occlusion as an inherent part of the inference problem, and aims to complete curves directly from partial observations without requiring physical interaction.

\paragraph{Diffusion Models for Occluded Objects.} 
In contrast to GANs \cite{goodfellow2014generative}, denoising diffusion probabilistic models (DDPMs) \cite{ho2020denoising}, have recently been adopted as a premier method for image generation, via their application in algorithms like Stable Diffusion \cite{rombach2022high}. This generative ability extends into the use case of inpainting \cite{lugmayr2022repaint}, in which a learning-based model generates pixels for a masked region, conditioned on known pixels from the unmasked portion. 

Building on this, Ozguroglu et al. present pix2gestalt \cite{ozguroglu2024pix2gestalt}, a promising method for generatively completing occluded objects via diffusion-based inpainting. However, the training and evaluation distribution for their method generally focuses on general rigid (or piecewise-rigid) objects, for which there is less pose uncertainty under occlusion when compared with curvilinear objects. In our work, we focus specifically on the curvilinear case, in which we hope that a specialized model may more effectively learn the distribution of curvilinear features.

Extending from the traditional DDPM approach to diffusion, our method instead leverages flow matching \cite{lipman2022flow}. While DDPMs require many sampling steps and involve complex training objectives, flow matching learns an explicit ODE path between noise and data, enabling more efficient sampling while maintaining generation quality. By adopting a flow matching, we instead opt to learn a more direct probability path between noise and data, enabling more efficient sampling while performing the same inpainting function.

\section{Problem Statement}
\label{sec:problem}
      Given an observed image $\mathbf{I}_{\text{obs}} \in \mathbb{R}^{H \times W \times 3}$ containing a curvilinear feature partially occluded by one or more obstacles, and optionally a binary mask $\mathbf{M} \in \{0,1\}^{H \times W}$ indicating occluded regions, we aim to recover the complete, unoccluded feature representation $\mathbf{I}_{\text{gt}}$. Formally, we seek a mapping $f_\theta: (\mathbf{I}_{\text{obs}}, \mathbf{M}) \mapsto \hat{\mathbf{I}}$ that minimizes the mean squared error (MSE) loss:
      \begin{equation}
      \label{eq:mse_loss}
      \mathcal{L} = \mathbb{E}_{\mathbf{I}_{\text{gt}}, \mathbf{M}} \left[ \alpha\|\hat{\mathbf{I}} \odot \mathbf{M} - \mathbf{I}_{\text{gt}} \odot \mathbf{M}\|_2^2 + \beta \|\hat{\mathbf{I}} - \mathbf{I}_{\text{gt}}\|_2^2 \right],
      \end{equation}
      where $\odot$ denotes element-wise multiplication and the coefficients $\alpha, \beta$ balance masked vs. global reconstruction loss.

      We consider two problem settings:
      \begin{enumerate}
      \item \textbf{Masked Inpainting}: $\mathbf{M}$ is provided as input. The task reduces to conditional generation where the model completes $\mathbf{I}_{\text{obs}}$ within $\mathbf{M}$ while preserving observed pixels.
      \item \textbf{Blind Completion}: $\mathbf{M}$ is unknown. The model must jointly estimate occlusion regions and complete the curvilinear feature, requiring implicit reasoning about discontinuity patterns.
      \end{enumerate}

      \section{Method}
      \label{sec:method}

\subsection{Architecture Overview}

ROCOCO comprises two sequentially trained UNet variants:

\begin{enumerate}
\item \textbf{Mask Predictor}: a UNet with 3-channel input, 64-dimensional hidden features and
      a 1-channel sigmoid output, trained with binary cross-entropy to predict
      $\hat{\mathbf{M}} = \sigma(g_\phi(\mathbf{I}_{\text{obs}}))$.
\item \textbf{Flow Matcher}: a time-conditioned UNet that predicts velocity fields for flow
      matching. It accepts the current state, the occlusion mask and a timestep, and outputs
      $\hat{v} \in \mathbb{R}^{H\times W\times 3}$.
\end{enumerate}

Both networks share core building blocks.

\paragraph{Residual blocks.} The base block implements
\begin{equation}
\mathbf{y} = \text{Conv}_{3\times3}\big(\text{SiLU}(\text{GN}(\text{Conv}_{3\times3}(
\text{SiLU}(\text{GN}(\mathbf{x})))))\big), \qquad
\hat{\mathbf{x}} = \mathbf{y} + \text{Conv}_{1\times1}(\mathbf{x}),
\end{equation}
with GroupNorm (32 groups) and SiLU activations. For time conditioning, a variant injects the
timestep embedding $\mathbf{e}_t$ between the two convolutions via a learned projection,
$\hat{\mathbf{x}} = \text{Conv}_2(\text{SiLU}(\text{GN}(\text{Conv}_1(\mathbf{x}) +
\text{Proj}(\mathbf{e}_t)))) + \text{Conv}_{\text{skip}}(\mathbf{x})$.

\paragraph{Attention.} To relate curve fragments on opposite sides of an occluder, the network
needs interactions that reach beyond the convolutional receptive field. We therefore apply
multi-head self-attention over the $HW$ \emph{spatial} positions of a feature map: the map is
flattened to a sequence of $HW$ tokens of width $C$, attended with four heads, and reshaped
back with a residual connection,
\begin{equation}
\text{Attn}(\mathbf{X}) = \text{Reshape}\big(\text{MHA}(\text{Flatten}(\mathbf{X}))\big) + \mathbf{X}.
\end{equation}
Because attention is quadratic in the number of tokens, applying it at full resolution is
infeasible at $128\times128$ ($16{,}384$ tokens). As in standard diffusion UNets we place it at
the coarsest ($\nicefrac{1}{4}$) level, where $32\times32 = 1024$ tokens make it cheap while
still spanning the whole image.

\textbf{Time embedding.} Timesteps $t \in [0,1]$ are embedded with a standard sinusoidal
positional encoding of dimension $d = 64$, with log-spaced frequencies, followed by a two-layer
MLP with a SiLU nonlinearity.

\subsection{Flow Matching Formulation}
\label{sec:flowmatching}

We adopt a flow matching objective \citep{lipman2022flow} with linear interpolation paths.
For each training sample $(\mathbf{x}_0, \mathbf{x}_{\text{obs}}, \mathbf{m})$:

\begin{enumerate}
\item Sample noise $\mathbf{x}_1 \sim \mathcal{N}(0,\mathbf{I})$ and $t \sim \mathcal{U}[0,1]$.
\item Form the interpolated state $\mathbf{x}_t = (1-t)\mathbf{x}_0 + t\mathbf{x}_1$.
\item Construct the conditional input by compositing the noisy state inside the occlusion with
      the observation outside it, and appending the mask as an explicit channel:
      $
      \mathbf{c}_t = \big[\; \mathbf{x}_t \odot \mathbf{m} + \mathbf{x}_{\text{obs}} \odot (1-\mathbf{m}) \;;\; \mathbf{m} \;\big].
      $
      
\item Predict the velocity $\hat{v} = v_\theta(\mathbf{c}_t, t)$.
\item Minimize the mask-area-normalized velocity error:
      $
      \mathcal{L} = \frac{\sum \mathbf{m} \odot \|\hat{v} - (\mathbf{x}_1 - \mathbf{x}_0)\|^2}{\sum \mathbf{m}}.
      $
\end{enumerate}


Sampling proceeds by Euler integration,
\begin{equation}
\mathbf{x}_{t-\Delta t} \leftarrow \mathbf{x}_t - \Delta t \cdot
v_\theta(\mathbf{c}_t, t) \odot \mathbf{m}, \qquad \Delta t = 1/T,
\end{equation}
with the observed pixels re-imposed after every step so that visible content is preserved
exactly. We study the choice of $T$ in \S\ref{sec:robustness}.

\subsection{Two-Stage Inference}
For blind completion, we cascade the mask predictor and flow matcher:
\begin{enumerate}
    \item Estimate occlusion mask: $\hat{\mathbf{m}} = \mathbb{I}[\sigma(g_\phi(\mathbf{I}_{\text{obs}})) > 0.05]$.
    \item Initialize masked noise state:
        $\mathbf{x}_t^{(0)} =
        \hat{\mathbf{m}} \odot \mathbf{x}_1
        +
        (1 - \hat{\mathbf{m}}) \odot \mathbf{I}_{\text{obs}},
        \quad \mathbf{x}_1 \sim \mathcal{N}(0, \mathbf{I}).$
    \item Iteratively update:
        $
        \mathbf{x}_t \leftarrow \mathbf{x}_t
        - \frac{1}{T} \;
        \hat{\mathbf{m}} \odot
        v_\theta\big(
            \hat{\mathbf{m}} \odot \mathbf{x}_t
            +
            (1-\hat{\mathbf{m}})\odot \mathbf{I}_{\text{obs}},\;
            t
        \big).
        $
    \item Output:
        $
        \hat{\mathbf{I}} =
        \hat{\mathbf{m}} \odot \mathbf{x}_0
        +
        (1-\hat{\mathbf{m}})\odot \mathbf{I}_{\text{obs}}.
        $
\end{enumerate}
This decoupling allows the flow matcher to focus on geometric completion while the mask predictor handles discontinuity detection.

\paragraph{Sampling multiple completions.} Because the completer is generative, repeated
sampling of the same input yields different plausible curves. Averaging $n$ independent
completions inside the occluded region is a cheap way to trade computation for accuracy, and we
report both the single-sample and averaged settings throughout.

\subsection{Additional Implementation Details}

All models use hidden dimension $64$, GroupNorm with 32 groups and SiLU activations, and are
trained with Adam. The flow matcher has $7.72$M parameters and the mask predictor $7.61$M.
Synthetic data is generated at $64\times64$ and the realistic datasets at $128\times128$.
Mask predictors are trained at learning rate $10^{-4}$ throughout. For the flow matcher we use
$10^{-3}$ on the wire dataset and $2\times10^{-4}$ on the synthetic and crack data, where we
found the higher rate to diverge late in training; all reported flow models are trained to a
stable optimum under their respective rate. Batch size is $64$ for the completers and $48$ for
the mask predictors.

\section{Experiments}
\label{sec:experiments}

\subsection{Datasets}

\paragraph{Synthetic data generation.}
\label{sec:datagen}
In order to train and evaluate our approach in a controlled manner, we procedurally generate ground truth images $\mathbf{I}_{\text{gt}}$, observed images $\mathbf{I}_{\text{obs}}$, and masks $\mathbf{M}$ by modeling the joint distribution \(p(\mathbf{I}_{\text{gt}}, \mathbf{I}_{\text{obs}}, \mathbf{M})\).
In our approach, control points \(\mathbf{P} \sim \mathcal{U}([0,W]\times[0,H])^N\), \( N\sim \mathcal{U}[4,6]\), are interpolated with a zero-smoothing B-spline to produce a curve \(\mathbf{C}\). Colors of the background, curve, and occlusion, \(c_{\text{bg}}, c_{\text{spline}}, c_{\text{occ}} \in [0,255]^3\), are sampled uniformly under constraints:
$
\|c_{\text{bg}} - c_{\text{spline}}\| > 100,\;
\|c_{\text{bg}} - c_{\text{occ}}\| > 60,\;
\|c_{\text{spline}} - c_{\text{occ}}\| > 60.
$
The ground-truth image \(\mathbf{I}_{\text{gt}}\) is obtained by rasterizing \(\mathbf{C}\) with random thickness on background \(c_{\text{bg}}\).

An occluder is generated by sampling a center \(\mathbf{c} \sim \text{Uniform}(\mathbf{C})\), angles \(\theta_i \sim \mathcal{U}(0,2\pi)\), a base radius \(R \sim \mathcal{U}(0.1W,0.25W)\), and radii \(r_i \sim \mathcal{U}(0.7R,1.3R)\), forming vertices \(\mathbf{q}_i = \mathbf{c} + r_i(\cos\theta_i,\sin\theta_i)\). Rasterization yields a mask \(\mathbf{M}\), and the observed image is:
$
\mathbf{I}_{\text{obs}} = \mathbf{I}_{\text{gt}} \odot (1 - \mathbf{M}) + c_{\text{occ}} \mathbf{M}.
$
For both training and evaluation, these synthetic samples are generated at runtime, which ensures that each new sample will be different from the rest. In our case, we elect to only employ sample \((\mathbf{I}_{\text{gt}}, \mathbf{I}_{\text{obs}}, \mathbf{M})\) if \(\mathbf{M}\) overlaps with \(\mathbf{C}\).

\paragraph{Realistic datasets.}
\label{sec:realdata}
To validate our approach on realistic data, we test our approach on two large datasets for the segmentation of curvilinear objects: A crack segmentation dataset aggregated from multiple public datasets \cite{zhang2016road, yang2019feature, eisenbach2017how, shi2016automatic, amhaz2016automatic, zou2012cracktree, shi2016automatic, cui2015pavement, liu2019deepcrack, yang2017deep, yang2017robotic} and an electric wire segmentation dataset provided by Zanella et al. \cite{zanella2021auto}. 
To modify these datasets for our own downstream prediction task, we resize all images to a size of $128 \times 128$ and introduce our own occluding polygons following the approach described in \hyperref[sec:datagen]{Synthetic Data Generation}, using the ground truth segmentation masks from the curve datasets to verify that each occluding polygon intersects with the curve in question. By applying our method to these types of realistic data, we demonstrate that our approach is robust and can be applied across relevant domains.
\subsection{Evaluation Metrics}

We employ the MSE of Eq.~\ref{eq:mse_loss} together with a more intuitively interpretable mean
$\ell_1$ loss (ML1L), the average $\ell_1$ distance between pixel colors over a selected
region. Pixel values are scaled from $[0,255]^3$ into the prediction range $[-1,1]^3$, so an
ML1L of $0.2$ corresponds to an average difference of $25.5$ in RGB per channel. We report both
metrics on the full image ($\alpha=0,\beta=1$ in Eq.~\ref{eq:mse_loss}) and on the occluded
region alone ($\alpha=1,\beta=0$), rescaling the latter so that it represents the average over
masked pixels only. Because pixel error over a largely static background can understate
differences in the completed structure, we additionally report Structural Similarity Index Measure (SSIM) over the occluded crop.

\subsection{Results and Ablations}
\label{sec:results}

\subsubsection{Synthetic Images}
\label{sec:synthres}

Table~\ref{tab:synth} reports the ablation of our approach on procedurally generated data.
\textbf{Naive UNet} predicts $\mathbf{I}_{\text{gt}}$ from $\mathbf{I}_{\text{obs}}$ in a single
forward pass. \textbf{Pure Flow} predicts $\mathbf{I}_{\text{gt}}$ by iterative flow prediction
from Gaussian noise, conditioned on $\mathbf{I}_{\text{obs}}$ through additional input channels
but never told where the occlusion is. \textbf{Flow + GT mask} performs flow-matching
inpainting with the ground-truth occlusion mask, and \textbf{Flow Blind} is the full two-stage
pipeline with the mask estimated from the observation.

\begin{table}[t]
\centering\small
\caption{Ablation on procedurally generated data, averaged over $500$ freshly sampled test
cases. Because samples are generated at run time, no test case is seen during training.
Subscript $M$ denotes the occluded region.}
\label{tab:synth}
\begin{tabular}{lccccc}
\toprule
Model & MSE$\downarrow$ & ML1L$\downarrow$ & MSE$_M\downarrow$ & ML1L$_M\downarrow$ & SSIM$_M\uparrow$ \\
\midrule
Naive UNet & 0.0026 & 0.0170 & 0.0548 & 0.1354 & 0.764 \\
Pure Flow & 0.0028 & 0.0061 & 0.0731 & 0.1666 & 0.708 \\
Flow + GT mask & \textbf{0.0018} & \textbf{0.0036} & \textbf{0.0398} & \textbf{0.0892} & \textbf{0.851} \\
Flow Blind (full pipeline) & 0.0019 & 0.0038 & 0.0427 & 0.0922 & 0.844 \\
\bottomrule
\end{tabular}

\end{table}

The ablation isolates where the benefit comes from. Flow-matching inpainting with the
occlusion mask reduces ML1L$_M$ to $0.089$ from the naive UNet's $0.135$, a $34\%$ reduction,
and raises SSIM$_M$ from $0.764$ to $0.851$ --- iterative generation conditioned on the visible
remainder recovers curvilinear structure that a single regression pass does not.

The mask supervision is essential to that gain rather than incidental to it. Pure Flow uses the
same flow-matching machinery and the same observation, but is never told where the occlusion
is; it reaches only $0.167$ ML1L$_M$, worse than the naive baseline. Generative modeling alone
is therefore not what helps: a model that must decide both what to draw and where to
draw it does worse than one that simply regresses the whole image. It is the combination of
flow matching applied to a localized region that produces the improvement, which is what
motivates the two-stage design.

Finally, the blind pipeline recovers almost all of the mask-supervised performance: $0.092$
against $0.089$ ML1L$_M$ and $0.844$ against $0.851$ SSIM$_M$, a gap of roughly $3\%$. The mask
predictor reaches an IoU of $0.977$ on this data, so replacing the ground-truth mask with an
estimated one costs very little, and the privileged information can be dispensed with almost
for free.

\begin{figure*}[h]
    \centering
    \includegraphics[width=\linewidth]{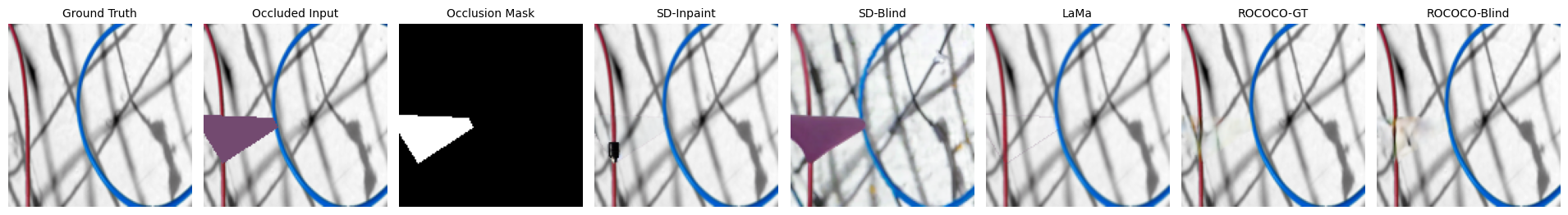}
    \includegraphics[width=\linewidth]{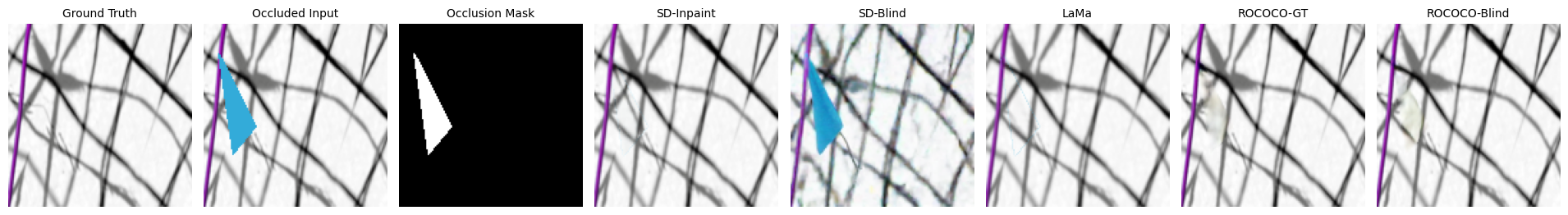}
    \caption{Qualitative comparison on the wire dataset for all baselines. Columns show: Ground Truth, Occluded Input, Occlusion Mask, SD-Inpaint, SD-Blind, LaMa, ROCOCO-GT, ROCOCO-Blind.}
    \label{fig:baseline_qualitative}
\end{figure*}

\subsubsection{Realistic Images}
\label{sec:realres}

\begin{figure}[t]
\centering
\includegraphics[width=0.42\linewidth]{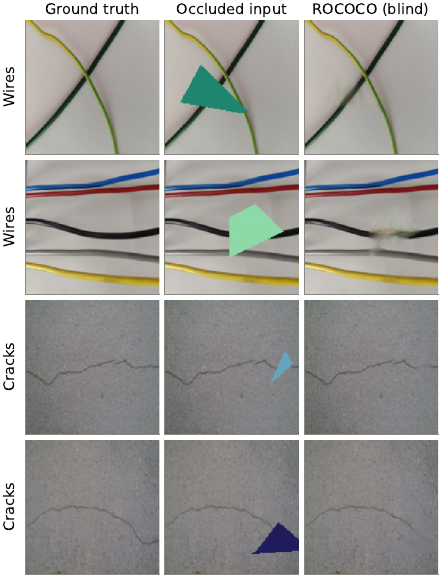}
\caption{Blind completions on held-out images from both datasets: the model receives only the
occluded input, and must localize the occluder and reconstruct the curvilinear structure
behind it without being told where it is.}
\label{fig:qualitative}
\end{figure}

We train and evaluate the full pipeline on the wire and crack segmentation datasets
(\S\ref{sec:realdata}) without privileged information about the occlusion mask.
Table~\ref{tab:real} reports performance on each held-out split, and
Figure~\ref{fig:qualitative} shows representative completions.

\begin{table}[t]
\centering\small
\caption{ROCOCO (blind) on the held-out splits of both realistic datasets.}
\label{tab:real}
\begin{tabular}{lccccc}
\toprule
Dataset & MSE$\downarrow$ & ML1L$\downarrow$ & MSE$_M\downarrow$ & ML1L$_M\downarrow$ & SSIM$_M\uparrow$ \\
\midrule
Wires \citep{zanella2021auto} & 0.0024 & 0.0064 & 0.0576 & 0.1670 & 0.655 \\
Cracks & 0.0022 & 0.0060 & 0.0696 & 0.1931 & 0.547 \\
\bottomrule
\end{tabular}

\end{table}

Both datasets are handled by the same pipeline with no architectural change. The mask predictor
localizes the occluder accurately in each case, getting IoU $0.944$ on wires and $0.939$ on cracks at
the operating threshold, so the blind pipeline retains almost all of the accuracy of the
mask-supervised one. Performance is comparable across the two domains despite their very
difference appearance: wires are high-contrast cables against smooth backdrops, while cracks are
thin, low-contrast fractures in textured concrete and asphalt. The crack images are also harder
in a specific way: their backgrounds are close to uniform, so more of the reconstruction error
is concentrated in the fine texture the model cannot predict, which is reflected in the lower
SSIM.

To confirm that these numbers reflect genuine completion rather than an easy task, we compare
against two trivial references. Leaving the occluder in place scores $0.565$ ML1L$_M$ on wires
and $0.577$ on cracks; filling the occluded region with the mean color of the visible image
scores $0.292$ and $0.217$. ROCOCO reaches $0.167$ and $0.193$, so it improves substantially on
both, and the margin is largest on wires, where there is real structure to recover.

\subsubsection{Comparison with Inpainting Baselines}
\label{sec:baselines}

We evaluate ROCOCO against off-the-shelf inpainting methods on the held-out wire split. For
methods that require an explicit occlusion mask we provide the ground truth; for blind variants
none is supplied. \textbf{SD-Inpaint} uses the \texttt{stable-diffusion-v1-5} inpainting
checkpoint via \textsc{Diffusers}: the $128\times128$ occluded image and mask are upsampled to
$512\times512$, the mask dilated by 3 pixels, with the prompt \textit{``wire(s) on a flat
background''}, 30 denoising steps and guidance $7.5$; the output is composited with the
observation and downsampled back. \textbf{SD-Blind} uses the standard checkpoint through the
image-to-image pipeline with no mask and a prompt instructing it to remove the occluder,
testing whether a general-purpose prior can localize and repair an occlusion unaided.
\textbf{LaMa} \citep{suvorov2021resolutionrobustlargemaskinpainting} is the \texttt{big-lama}
checkpoint, run at $256\times256$ with a 3-pixel dilated mask and composited back. ROCOCO uses
$T=5$ integration steps, reported both with a single sample and averaged over $n=8$, as suggested earlier
(\S\ref{sec:robustness}).
See Table~\ref{tab:baselines} for quantitative and Figure~\ref{fig:baseline_qualitative} for qualitative comparison.

\begin{table}[t]
\centering\small
\caption{Comparison against off-the-shelf inpainting on the held-out wire split ($62$ images).
``Mask'' indicates whether the method is given the ground-truth occlusion mask; ROCOCO's blind
rows estimate it from the observation alone. Subscript $M$ denotes the occluded region.}
\label{tab:baselines}
\begin{tabular}{llcccccc}
\toprule
Method & Mask & Params & MSE$\downarrow$ & ML1L$\downarrow$ & MSE$_M\downarrow$ & ML1L$_M\downarrow$ & SSIM$_M\uparrow$ \\
\midrule
SD-Blind & no & 860M & 0.0474 & 0.1407 & 0.4416 & 0.5411 & 0.258 \\
SD-Inpaint & GT & 860M & 0.0041 & 0.0199 & 0.0895 & 0.1923 & 0.634 \\
LaMa & GT & 51M & 0.0037 & 0.0297 & 0.0560 & \textbf{0.1468} & \textbf{0.695} \\
\midrule
ROCOCO & GT & \textbf{7.7}M & 0.0032 & 0.0073 & 0.0782 & 0.1955 & 0.596 \\
ROCOCO ($n{=}8$) & GT & \textbf{7.7}M & \textbf{0.0022} & \textbf{0.0060} & \textbf{0.0540} & 0.1615 & 0.664 \\
ROCOCO & no & \textbf{7.7}M & 0.0035 & 0.0078 & 0.0831 & 0.2012 & 0.586 \\
ROCOCO ($n{=}8$) & no & \textbf{7.7}M & 0.0024 & 0.0064 & 0.0576 & 0.1670 & 0.655 \\
\bottomrule
\end{tabular}

\end{table}

\textbf{When no occlusion mask is available ROCOCO is the strongest method by a wide
margin}. The only other mask-free method, SD-Blind, reaches $0.541$ ML1L$_M$ against ROCOCO's
$0.167$ --- a factor of $3.2$ --- and $0.258$ SSIM against $0.655$. A general-purpose
image-to-image prior cannot reliably localize a geometric occluder and repair the structure
behind it, whereas our lightweight two-stage pipeline can. This is the setting that matters for the
motivating application, since an occluder is not annotated at test time.

\textbf{When a mask is available ROCOCO remains competitive with dedicated
mask-supervised inpainting at a fraction of the size}. It obtains the best full-image MSE
($0.0022$ against LaMa's $0.0037$), the best full-image ML1L ($0.0060$ against $0.0297$) and
the best masked MSE ($0.0540$ against $0.0560$), and is second to LaMa on masked ML1L ($0.1615$
against $0.1468$) and SSIM ($0.664$ against $0.695$), while comfortably clears SD-Inpaint on
every metric. It does so with a $7.7$M-parameter completer, against $51$M for LaMa and an
$860$M UNet for Stable Diffusion (i.e., roughly a seventh and a hundredth of the parameters
respectively). LaMa retains an advantage in fine detail inside the occlusion, which is
consistent with its much larger capacity and effective receptive field, but it cannot operate
without a mask at all.

Notably, moving from the ground-truth mask to the predicted one costs very little: ML1L$_M$
rises only from $0.1615$ to $0.1670$ ($3.4\%$ relative), because the mask predictor is accurate
(IoU $0.944$ at the operating threshold). The privileged information the other mask-supervised
methods depend on is, for our pipeline, almost free to recover.

\subsection{Robustness and Inference Cost}
\label{sec:robustness}

\begin{figure}[t]
\centering
\includegraphics[width=0.80\linewidth]{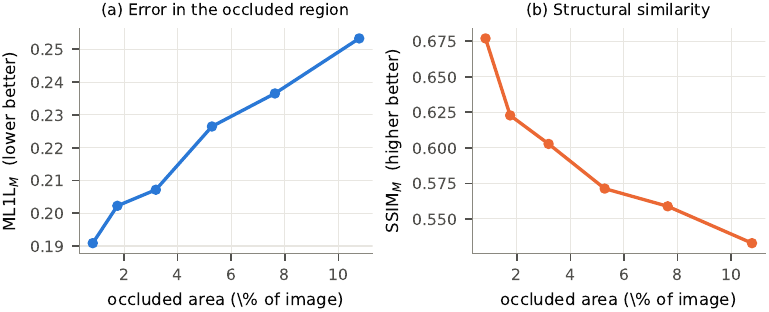}
\caption{Completion quality against occluder size on the held-out wire split. Degradation is
graceful: the occluded-region error rises by about a third as the occluded area grows more than
tenfold.}
\label{fig:size}
\end{figure}

\textbf{Occluder size.} We regenerate the evaluation set with the occluder radius swept from
$6\%$ to $35\%$ of image width, giving occluded areas from $0.8\%$ to $10.3\%$
(Figure~\ref{fig:size}). Completion degrades monotonically as the gap widens, as expected, with
ML1L$_M$ rising from $0.191$ to $0.253$ and SSIM$_M$ falling from $0.677$ to $0.533$ across the
range, which is a graceful degradation rather than a collapse, even when the occluder hides a tenth
of the image.

\begin{figure}[t]
\centering
\includegraphics[width=0.80\linewidth]{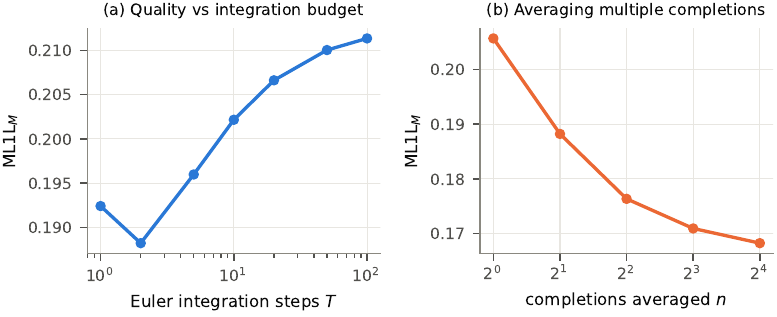}
\caption{Inference budget. Quality peaks at $T\approx2$ integration steps rather than the tens
typically used for iterative samplers, and averaging independent completions is a more
productive use of the same compute.}
\label{fig:budget}
\end{figure}

\textbf{Integration steps and sample averaging.} Sampling cost is linear in the number of
Euler steps $T$. We find that quality peaks at $T\approx2$ and then degrades slowly:
ML1L$_M$ rises from $0.188$ at $T=2$ to $0.210$ at $T=50$ and $0.211$ at $T=100$
(Figure~\ref{fig:budget}). With a linear probability path the velocity field is close to
constant along the trajectory, so a coarse integrator suffices and further steps mostly simply compound
the model's own error. Because the completer is generative, averaging $n$ independent
completions seem to be a more productive use of the same budget: at $T=5$, ML1L$_M$ falls from $0.206$
at $n=1$ to $0.171$ at $n=8$ and $0.168$ at $n=16$, so the return has largely saturated by
$n=8$. We therefore use $T=5$ and $n=8$, which costs $40$ network evaluations of a
$7.7$M-parameter model.

\section{Conclusion}
\label{sec:conclusion}
ROCOCO presents an efficient method for the completion of occluded curvilinear objects. A
comparison against a one-shot UNet demonstrates that a flow-matching formulation provides
substantial benefits in prediction quality; ablating the mask information shows that
supervising the occluded region specifically improves reconstruction of the curve rather than
merely the background. By pairing the completer with a mask predictor, we obtain that
supervision without privileged information about the occluded region, which the mask-supervised
baselines cannot do. We validate our method both on programmatic spline-image data and on realistic curve images from large public datasets.

Against off-the-shelf inpainting the picture is consistent: with no occlusion mask ROCOCO is far
stronger than the only comparable alternative, and with one it remains competitive with
dedicated inpainting models, achieving best on full-image error and second on the occluded region,
while using roughly a seventh of LaMa's parameters and a hundredth of Stable Diffusion's. Moving from
a ground-truth mask to a predicted one costs only a few percent.

\textbf{Limitations and future work.} 
At this stage, the selection of particular hyperparameters, randomization ranges, and thresholds has not been explored exhaustively. 
The same can be said for the distribution of data scenarios, such as multiple curves, intersections, and greater variability in occlusions. Future work may want to explore variability in color across a curved object (such as that produced by shadows), or apply this method in conjunction with curve tracing techniques that benefit from unoccluded features to produce complete traces.
Overall we
believe our method provides a lightweight yet effective way to account for occlusion in the
perception of curvilinear objects, and look forward to the benefits it offers in relevant
downstream tasks.


\bibliographystyle{plain}
\bibliography{references}

\clearpage
\suppressfloats
\begin{figure*}
    {\Large\bfseries Appendix\par\medskip}
    {\large\bfseries Additional Figures\par\medskip}
    \label{sec:additional_figures}
\end{figure*}
\begin{figure*}[h!]
    \centering
    \includegraphics[width=0.80\linewidth]{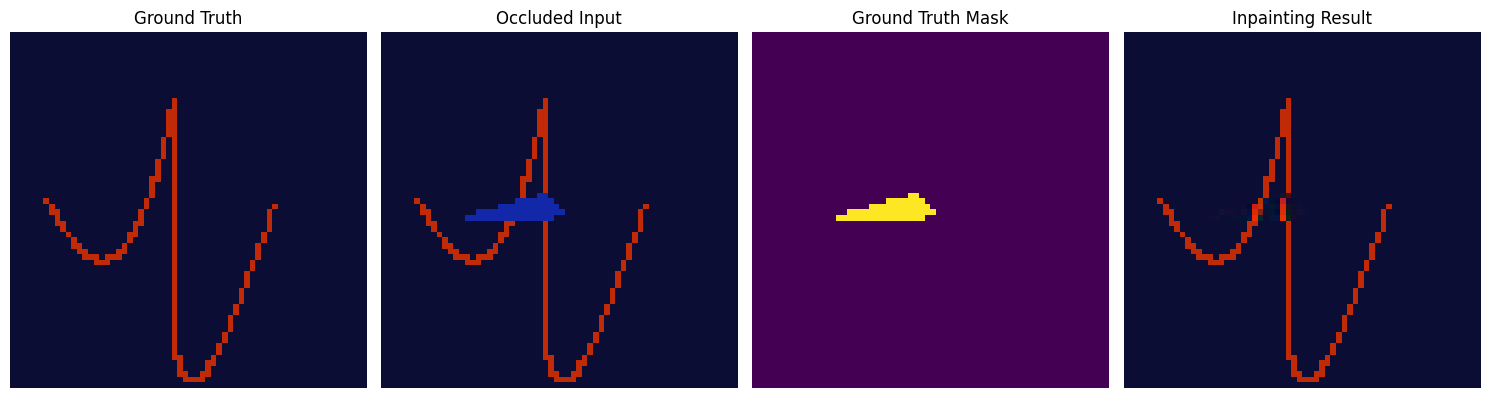}
    \caption{Example Generation Result: Inpainting with Ground Truth Mask}
\end{figure*}

\begin{figure*}[h!]
    \centering
    \includegraphics[width=0.80\linewidth]{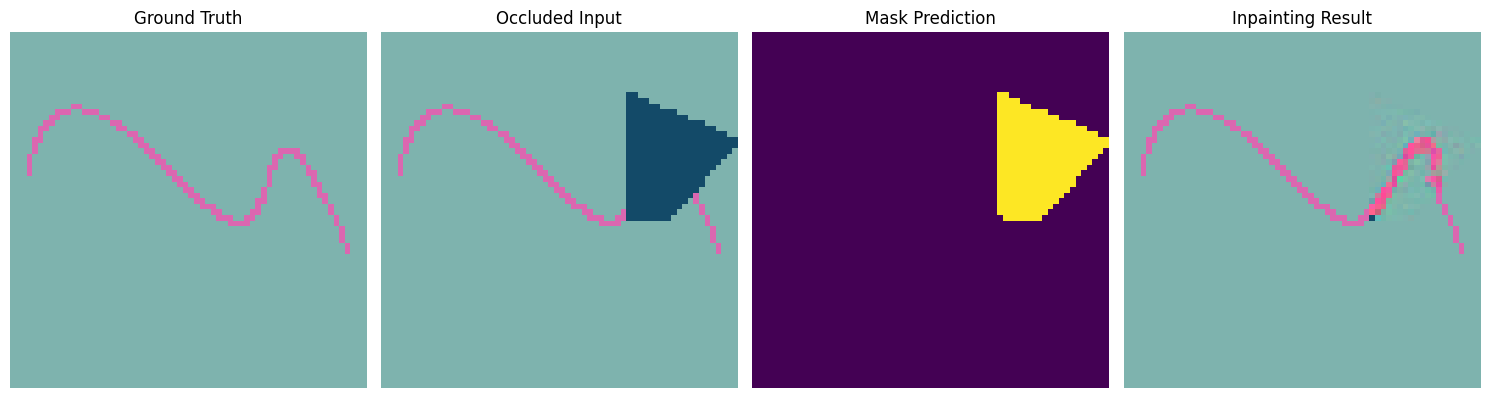}
    \caption{Example Generation Result: Inpainting with Mask Prediction}
\end{figure*}

\begin{figure*}[h!]
    \centering
    \includegraphics[width=0.80\linewidth]{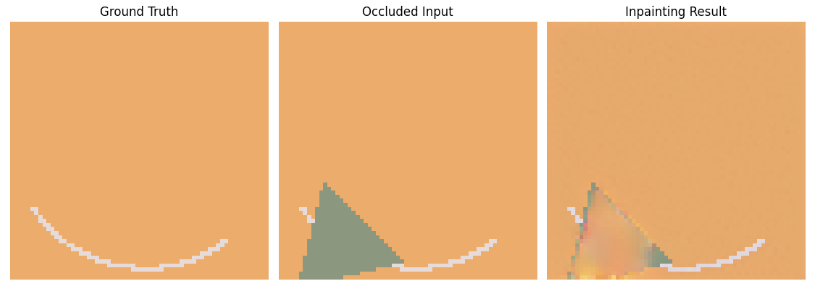}
    \caption{Example Generation Result: Pure Flow Model}
\end{figure*}

\begin{figure*}
    \centering
    \includegraphics[width=0.80\linewidth]{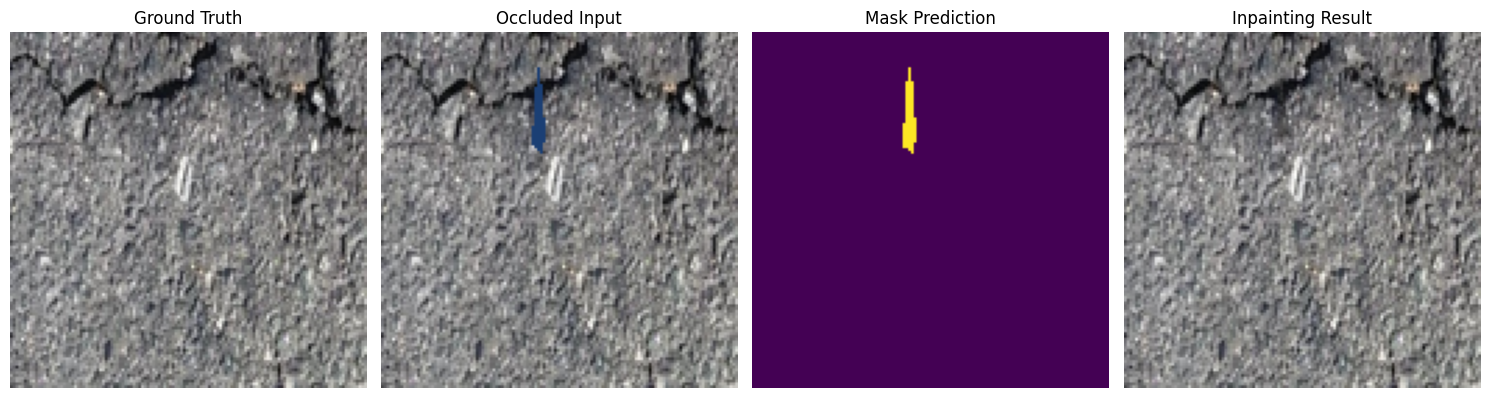}
    \includegraphics[width=0.80\linewidth]{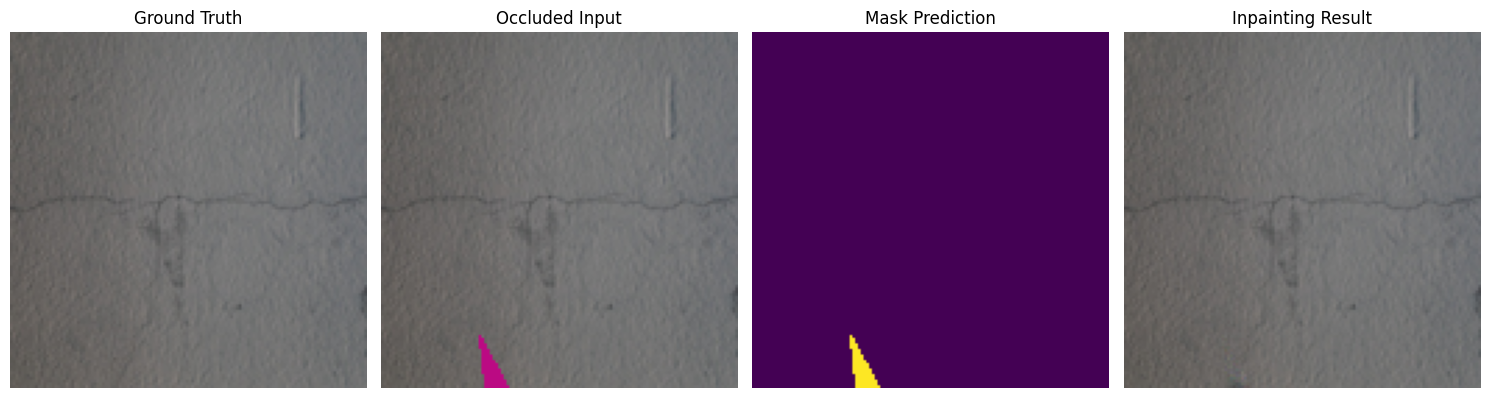}
    \includegraphics[width=0.80\linewidth]{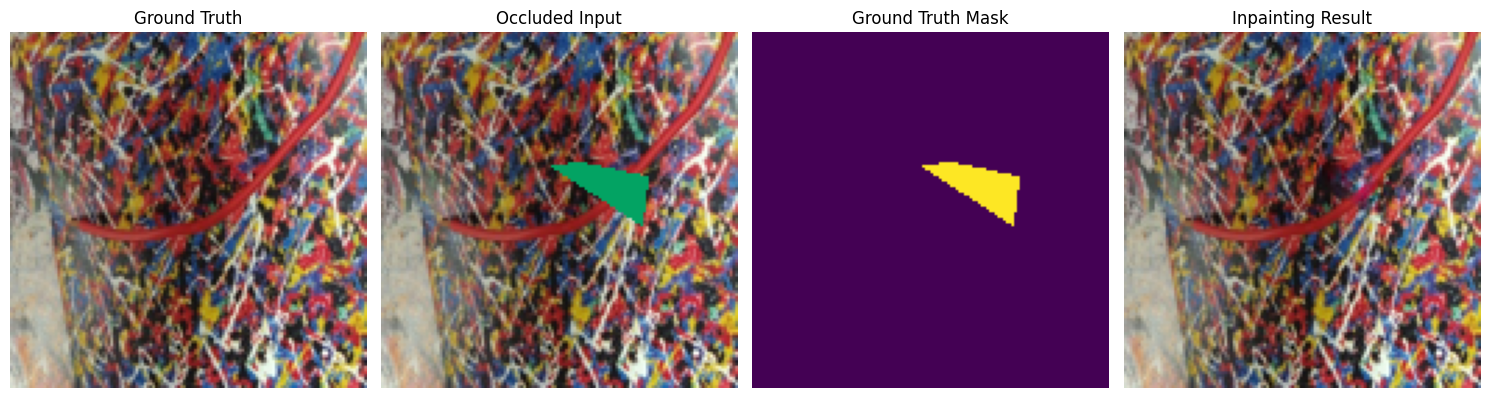}
    \includegraphics[width=0.80\linewidth]{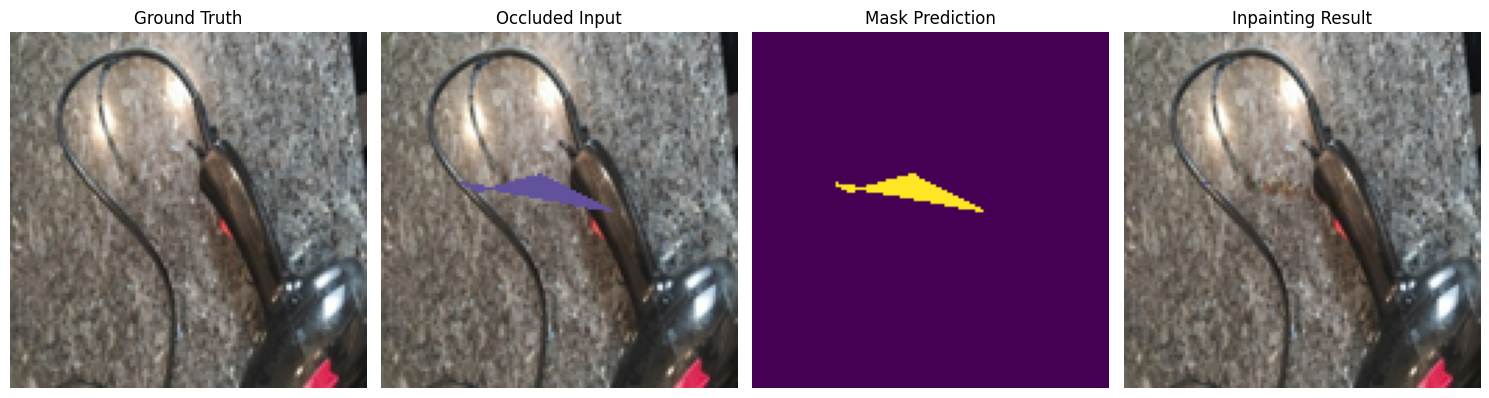}
    \caption{Example Generations: ROCOCO on cracks and wires}
\end{figure*}

\begin{figure*}
    \centering
    \includegraphics[width=\linewidth]{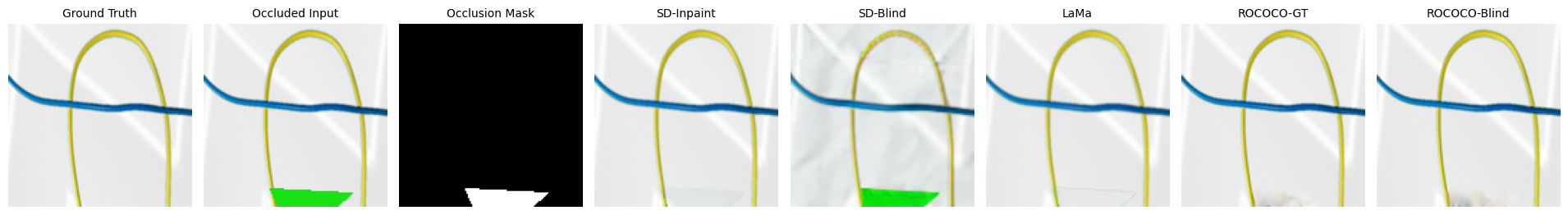}
    \caption{Example Baseline Comparison Result on the Wires Dataset}
\end{figure*}


\begin{figure*}
    \centering
    \includegraphics[width=0.60\linewidth]{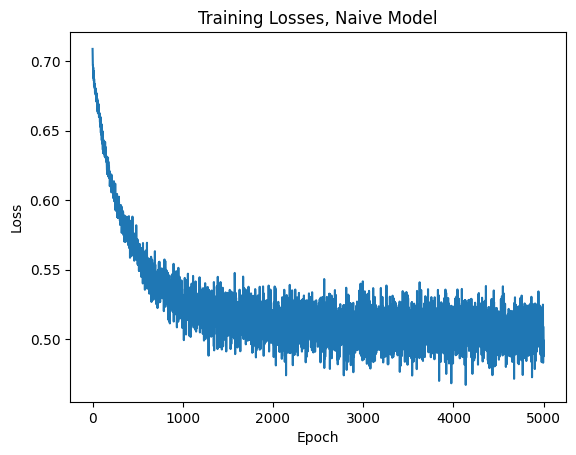}
    \caption{Training Losses, Naive UNet}
\end{figure*}

\begin{figure*}
    \centering
    \includegraphics[width=0.60\linewidth]{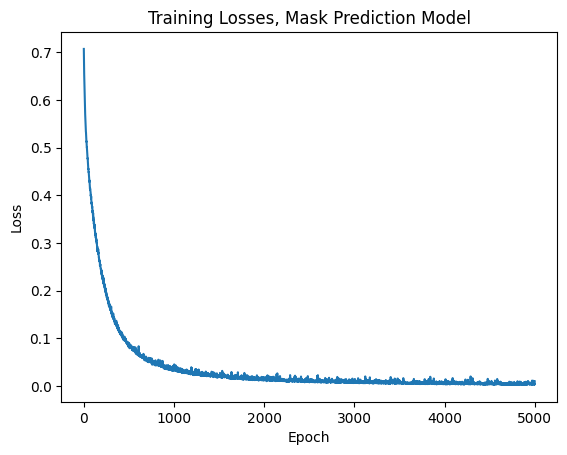}
    \caption{Training Losses, Mask Prediction Model}
\end{figure*}

\begin{figure*}
    \centering
    \includegraphics[width=0.60\linewidth]{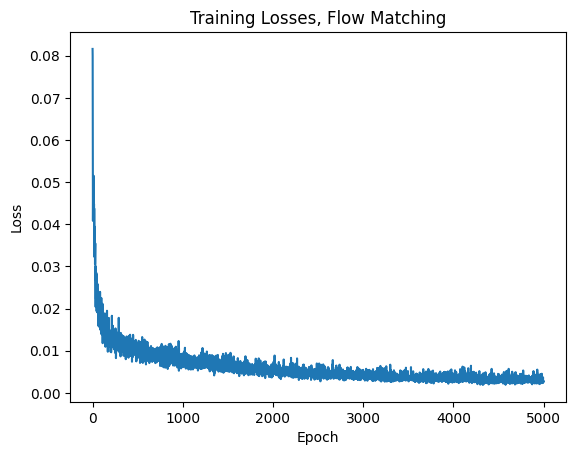}
    \caption{Training Losses, Flow Matching Inpainting Model}
\end{figure*}

\end{document}